\documentclass[a4paper,twoside]{article}

\usepackage{graphicx}
\usepackage{float}
\usepackage{multirow}
\usepackage{amsmath,amssymb} 
\usepackage{hyperref}
\usepackage{amsmath}
\usepackage{amssymb}
\usepackage{subcaption}
\usepackage{color,soul}
\usepackage{multirow}
\usepackage{rotating}

\usepackage{epsfig}
\usepackage{calc}
\usepackage{amssymb}
\usepackage{amstext}
\usepackage{amsmath}
\usepackage{amsthm}
\usepackage{multicol}
\usepackage{pslatex}
\usepackage{apalike}
\usepackage{SCITEPRESS}     

\begin{document}

\title{Multi-stream CNN based Video Semantic Segmentation \\
for Automated Driving}

\author{\authorname{Ganesh Sistu\sup{1}, Sumanth Chennupati\sup{2} and  Senthil Yogamani\sup{1}}
\affiliation{\sup{1}Valeo Vision Systems, Ireland}
\affiliation{\sup{2}Valeo Troy, United States}
\email{\{ganesh.sistu,sumanth.chennupati,senthil.yogamani\}@valeo.com}
}


\abstract{Majority of semantic segmentation algorithms operate on a single frame even in the case of videos. In this work, the goal is to exploit temporal information within the algorithm model for leveraging motion cues and temporal consistency. We propose two simple high-level architectures based on Recurrent FCN (RFCN) and Multi-Stream FCN (MSFCN) networks. In case of RFCN, a recurrent network namely LSTM is inserted between the encoder and decoder. MSFCN combines the encoders of different frames into a fused encoder via 1x1 channel-wise convolution. We use a ResNet50 network as the baseline encoder and construct three networks namely MSFCN of order 2 \& 3 and RFCN of order 2. MSFCN-3 produces the best results with an accuracy improvement of 9\% and 15\% for Highway and New York-like city scenarios in the SYNTHIA-CVPR'16 dataset using mean IoU metric.  MSFCN-3 also produced 11\% and 6\% for SegTrack V2 and DAVIS datasets over the baseline FCN network. We also designed an efficient version of MSFCN-2 and RFCN-2 using weight sharing among the two encoders. The efficient MSFCN-2 provided an improvement of 11\% and 5\% for KITTI and SYNTHIA  with negligible increase in computational complexity compared to the baseline version.}


\onecolumn \maketitle \normalsize \vfill

\section{Introduction}

Semantic segmentation provides complete semantic scene understanding wherein each pixel in an image is assigned a class label. It has applications in various fields including automated driving \cite{horgan2015vision} \cite{heimberger2017computer}, augmented reality and medical image processing. Our work is focused on semantic segmentation applied to automated driving which is discussed in detail in the survey paper \cite{siam2017deep}. Recently, this algorithm has matured in accuracy which is sufficient for commercial deployment due to advancements in deep learning. Most of the standard architectures make use of a single frame even when the algorithm is run on a video sequence. Efficient real-time semantic segmentation architectures are an important aspect for automated driving \cite{siam2018rtseg}. For automated driving videos, there is a strong temporal continuity and constant ego-motion of the camera which can be exploited within the semantic segmentation model. This inspired us to explore temporal based video semantic segmentation. 
This paper is an extension of our previous work on RFCN \cite{siam2017recurrent}. 

\begin{figure*}[!htbp]
     \centering
    \includegraphics[width=\textwidth]{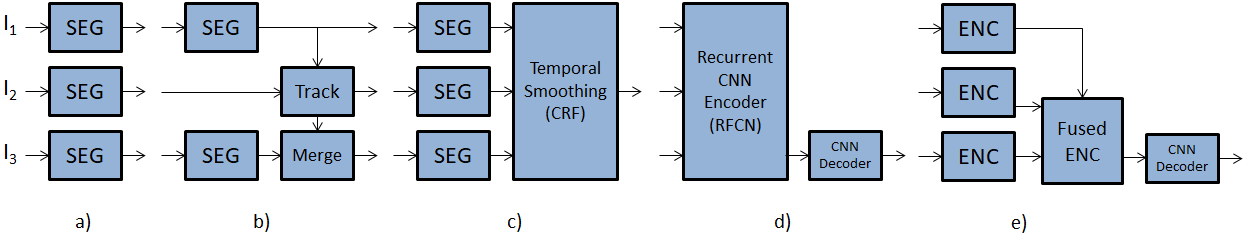}
    \caption{Comparison of different approaches to extend semantic segmentation to videos - a) Frame-level output b) Detect and track c) Temporal post processing d) Recurrent encoder model and e) Fused multi-stream encoder model.}
    \label{fig:video-extn}
\end{figure*}

In this paper, we propose two types of architectures namely Recurrent FCN (RFCN) and Multi-Stream FCN (MSFCN) inspired by FCN and Long short-term memory (LSTM) networks. Multi-Stream Architectures were first introduced in \cite{simonyan2014two} in which a two stream CNN was proposed for action recognition. They were also successfully used for other applications like Optical Flow \cite{ilg2016flownet}, moving object detection \cite{siam2017modnet} and depth estimation \cite{ummenhofer2016demon}. However, this has not been explored for semantic segmentation using consecutive video frames to the best of our knowledge. The main motivation is to leverage temporal continuity in video streams. In RFCN, we temporally processed FCN encoders using LSTM network. In MSFCN architecture, we combine the encoder of current and  previous frames to produce a new fused encoder of same feature map dimension. This would enable keeping the same decoder. \\ 

\noindent The list of contributions include:  
\begin{itemize}
    \item Design of  RFCN \& MSFCN architectures that extends semantic segmentation models for videos.
    \item Exploration of weight sharing among encoders for computational efficiency.
    \item Implementation of an end-to-end training method for spatio-temporal video segmentation.
    \item Detailed experimental analysis of video semantic segmentation with automated driving datasets KITTI \& SYNTHIA and binary video segmentation with DAVIS \& SegTrack V2 datasets.
\end{itemize}


The rest of the paper is structured as follows. Section \ref{video-extn} discusses different approaches for extending semantic segmentation to videos.  Section \ref{method} explains the different multi-stream architectures designed in this work. Experimental setup and results are discussed in section \ref{experiments}. Finally, section \ref{conclusion} provides the conclusion and future work.
 
\section{Extending Semantic segmentation to Videos} \label{video-extn}

In this section, we provide motivation for incorporating temporal models in automated driving and explain different high level methods to accomplish the same. Motion is a dominant cue in automated driving due to persistent motion of the vehicle on which the camera is mounted. The objects of interest in automotive are split into static infrastructure like road, traffic signs, etc and dynamic objects which are interacting like vehicles and pedestrians. The main challenges are posed due to the uncertain behavior of dynamic objects. Dense optical flow is commonly used to detect moving objects purely based on motion cues. Recently, HD maps is becoming a commonly used cue which enables detection of static infrastructure which is previously mapped and encoded. In this work, we explore the usage of temporal continuity to improve accuracy by implicitly learning motion cues and tracking. We discuss the various types of temporal models in Fig \ref{fig:video-extn} which illustrates the different ways to extend image based segmentation algorithm to videos. \\

\noindent \textbf{Single frame baseline:} Fig \ref{fig:video-extn} (a) illustrates the typical way the detector is run every frame independently. This would be the reference baseline for comparing accuracy of improvements by other methods. \\

\noindent \textbf{Detect and Track approach:}
The premise of this approach is to leverage the previously obtained estimate of semantic segmentation as the next frame has only incrementally changed. This can reduce the computational complexity significantly as a lighter model can be employed to refine the previous semantic segmentation output for the current frame. The high level block diagram is illustrated in Fig\ref{fig:video-extn} (b). This approach has been successfully used for detection of bounding box objects where tracking could even help when detector fails in certain frames. However, it is difficult to model it for semantic segmentation as the output representation is quite  complex and it is challenging to handle appearance of new regions in the next frame. \\

\noindent \textbf{Temporal post processing:}
The third approach is to use a post-processing filter on output estimates to smooth out the noise. Probabilistic Graphical Models (PGM) like Conditional Random Fields (CRF) are commonly used to accomplish this. The block diagram of this method is shown in Fig \ref{fig:video-extn} (c) where recurrence is built on the output. This step is computationally complex because the recurrence operation is on the image dimension which is large. \\

\noindent \textbf{Recurrent encoder model:}
 In this approach, the intermediate feature maps from the encoders are fed into a recurrent unit. The recurrent unit in the network can be an RNN, LSTM or a GRU. Then the resulting features are fed to a decoder which outputs semantic labels. In Fig \ref{fig:a}, the ResNet50 encoder conv5 layer features from consecutive image streams are passed as temporal features for LSTM network. While conv4 and conv3 layer features can also be processed via the LSTM layer, the conv4 and conv3 features from two stream are concatenated followed by a convolution layer to keep the architecture simple and memory efficient. \\

\noindent \textbf{Fused multi-stream encoder model:}
This method can be seen as a special case of Recurrent model in some sense. But the perspective of multi-stream encoder will enable the design of new architectures. As this is the main contribution of this work,  we will describe it in more detail in next section.

\section{Proposed CNN Architectures} \label{method}

\begin{figure} 
\centering     
\begin{subfigure}{0.5\textwidth} 
    \includegraphics[width=7cm,height=2cm]{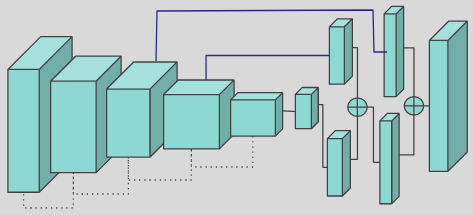}
    \caption{\textcolor{black}{FCN: Single Encoder Baseline\newline}} 
    \label{fig:a}
\end{subfigure}%

\begin{subfigure}{0.5\textwidth} 
    \includegraphics[width=7cm,height=4cm]{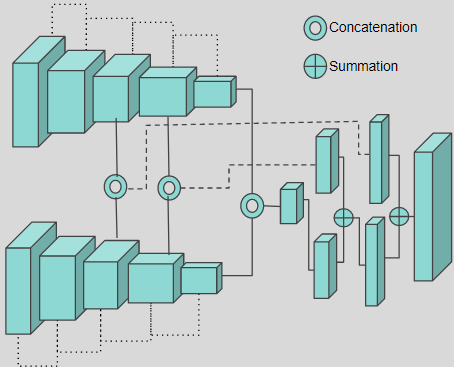}
    \caption{\textcolor{black}{MSFCN-2: Two stream fusion architecture\newline}}   
    \label{fig:b}
\end{subfigure}%

\begin{subfigure}{0.5\textwidth} 
    \includegraphics[width=7cm,height=3.5cm]{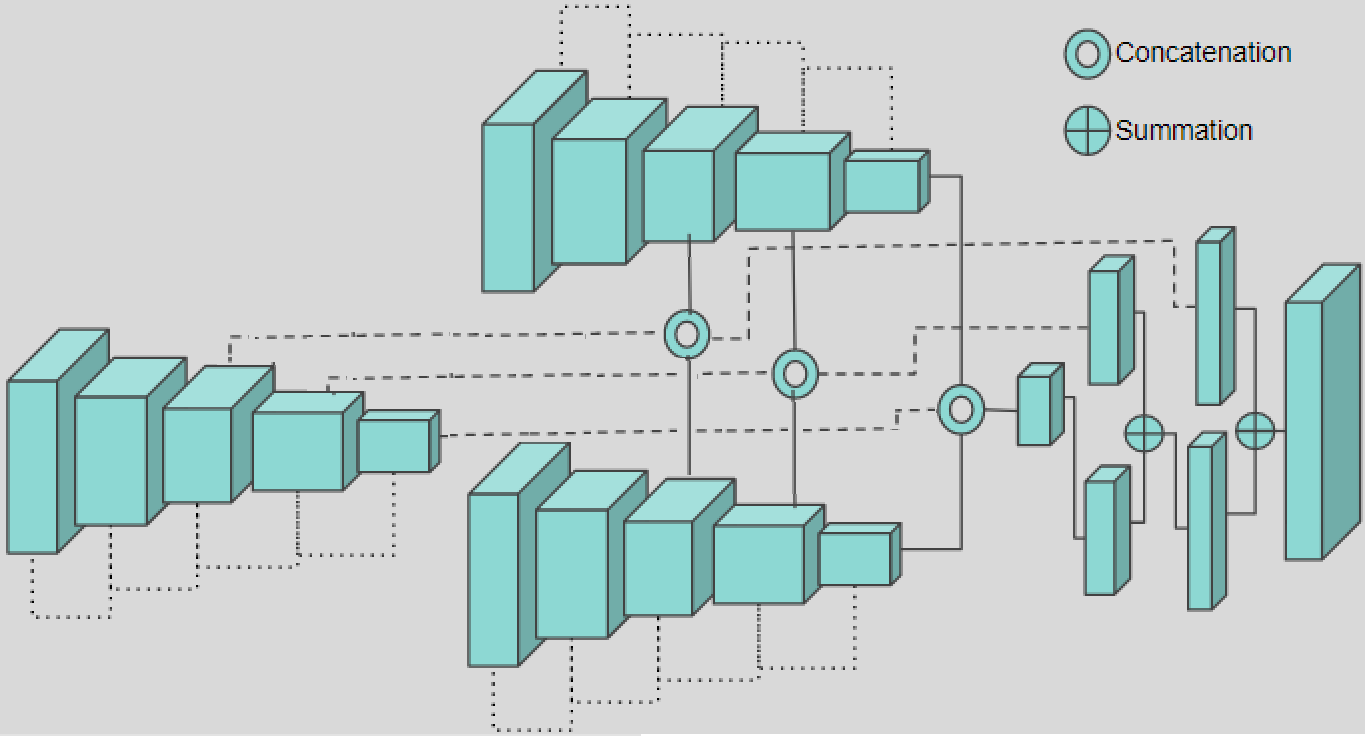}
    \caption{\textcolor{black}{MSFCN-3: Three stream fusion architecture\newline}}    
    \label{fig:c}
\end{subfigure}%

\begin{subfigure}{0.5\textwidth} 
    \includegraphics[width=7cm,height=4cm]{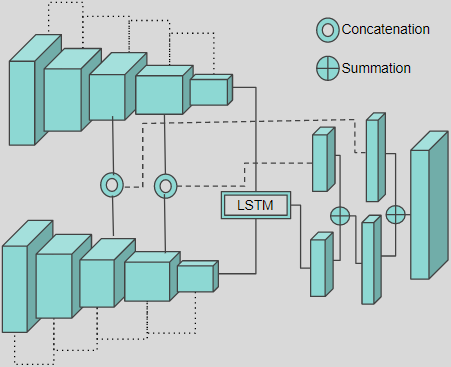}
    \caption{\textcolor{black}{RMSFCN-2: Two stream LSTM architecture\newline}}   
    \label{fig:d}
\end{subfigure}%

\caption{Four types of architectures constructed and tested in the paper. (a) Single stream baseline, (b) Two stream fusion architecture,  (c) Three stream fusion architecture and (d) Two stream LSTM architecture. }
\end{figure}

In this section, we discuss the details of the proposed multi-stream networks shown in Fig \ref{fig:b}, \ref{fig:c} \& \ref{fig:d}. Multi stream fused architectures (MSFCN-2 \& MSFCN-3) concatenate the output from each encoder and fuse them via 1x1 channel-wise convolutions to obtain a fused encoder which is then fed to the decoder. Recurrent based architecture (RFCN) uses an LSTM unit to feed the decoder. \\

\noindent \textbf{Single stream architecture:}
A fully convolution network (FCN) shown in Fig \ref{fig:a} is inspired from \cite{long2015fully} is used as the baseline architecture. We used ResNet50 \cite{HeZRS15} as the encoder and  conventional up-sampling with skip-connections to predict pixel wise labels. Initializing model weights by pre-trained ResNet50 weights, alleviates over-fitting problems as these weights are the result of training on a much larger dataset namely ImageNet. \\

\noindent \textbf{Multi-stream fused architectures:}
Multi-Stream FCN architecture is illustrated in Fig  \ref{fig:b} \& \ref{fig:c}. We used multiple ResNet50 encoders to construct the multi-stream architectures. Consecutive input frames are processed by multiple ResNet50 encoders independently. The intermediate feature maps obtained at 3 different stages (conv3, conv4 and conv5) of encoder are concatenated and added to the up-sampling layers of the decoder. MSFCN-2 is constructed using 2 encoders while MSFCN-3 uses 3 encoders. A channel-wise 1x1 convolution is applied to fuse the multiple encoder streams into a single one of the same dimension. This will enable the usage of the same decoder. \\

\begin{table*}[h]
\tiny
\centering
\caption{Semantic Segmentation Results on SYNTHIA Sequences. We split the test sequences into two parts, one is Highway for high speeds and the other is City for medium speeds.}
\label{table:results}
\begin{tabular}{|l|l|c|c|c|c|c|c|c|c|c|c|}
\hline
Dataset                  & Architecture & Mean IoU  & Sky  & Building & Road & Sidewalk & Fence & Vegetation & Pole & Car  & Lane \\ \hline
\multirow{4}{*}{Highway} & FCN          & 85.42 & 0.91 & 0.67     & 0.89 & 0.02     & 0.71  & 0.79       & 0.01 & 0.81 & \textbf{0.72}     \\ \cline{2-12} 
                         & MSFCN-2      & 93.44 & 0.92 & 0.66     & 0.94 & 0.28     & 0.85  & 0.78       & 0.11 & 0.82 & 0.71  \\ \cline{2-12} 
                         & RFCN-2       & 94.17 & \textbf{0.93} & \textbf{0.71}     & 0.95 & \textbf{0.31}     & 0.82  & \textbf{0.83}       & \textbf{0.13} & \textbf{0.87} & 0.7          \\ \cline{2-12} 
                         & MSFCN-3      & \textbf{94.38} & \textbf{0.93} & 0.69     & \textbf{0.96} & \textbf{0.31}     & \textbf{0.87}  & 0.81       & 0.12 & \textbf{0.87} & \textbf{0.72} \\ \hline
\multirow{4}{*}{City}    & FCN     & 73.88 & \textbf{0.94} & \textbf{0.94}     & 0.72 & 0.78     & 0.34  & 0.54       & 0    & 0.69 & 0.56          \\ \cline{2-12} 
                         & MSFCN-2 & 87.77 & 0.87 & \textbf{0.94}     & 0.84 & \textbf{0.83}     & \textbf{0.68}  & 0.64       & 0    & \textbf{0.8}  & \textbf{0.8}           \\ \cline{2-12} 
                         & RFCN-2  & 88.24 & 0.91 & 0.92     & \textbf{0.87} & 0.78     & 0.56  & \textbf{0.67}       & 0    & \textbf{0.8}  & 0.74          \\ \cline{2-12} 
                         & MSFCN-3 & \textbf{88.89} & 0.88 & 0.89     & 0.86 & 0.74     & 0.64  & 0.53       & 0    & 0.71 & 0.72          \\ \hline 
\end{tabular}

\bigskip
\centering
\caption{Semantic Segmentation Results on KITTI Video Sequence.}
\label{table:Kitti}

\begin{tabular}{|l|l|l|l|l|l|l|l|l|l|l|}

\hline

Architecture         & NumParams & Mean IoU         & Sky              & Building         & Road             & Sidewalk         & Fence            & Vegetation       & Car              & Sign             \\ \hline
FCN                  & 23,668,680 & 74.00          & 46.18          & 86.50          & 80.60          & 69.10          & 37.25          & 81.94          & 74.35          & 35.11          \\ \hline
MSFCN-2 (shared weights) & 23,715,272 & 85.31          & 47.89          & 91.08          & \textbf{97.58} & 88.02          & \textbf{62.60} & 92.01          & \textbf{90.26} & 58.11          \\ \hline
RFCN-2 (shared weights)  & 31,847,828 & 84.19          & \textbf{50.20} & \textbf{93.74} & 94.90          & 88.17          & 59.73          & 87.73          & 87.66          & 55.55          \\ \hline
MSFCN-2              & 47,302,984 & \textbf{85.47} & 48.72          & 92.29          & 96.36          & 90.21          & 59.60          & \textbf{92.43} & 89.27          & \textbf{70.47} \\ \hline
RFCN-2               & 55,435,540 & 83.38          & 44.80          & 92.84          & 91.77          & \textbf{91.67} & 58.53          & 86.01          & 87.25          & 52.87          \\ \hline
\end{tabular}

\bigskip
\centering
\caption{Semantic Segmentation Results on SYNTHIA Video Sequence.}
\label{table:syncity}
\begin{tabular}{|l|l|l|l|l|l|l|l|l|l|l|l|l|l|}
\hline
Architecture         & Mean IoU       & Sky            & Building       & Road           & Sidewalk       & Fence          & Vegetation    & Pole           & Car            & Sign           & Pedestrain     & Cyclist       & Lane           \\ \hline
FCN                  & 84.08          & 97.2           & 92.97          & 87.74          & 81.58          & 34.44          & 62            & 1.87           & 72.75          & 0.21           & 0.01           & 0.33          & 93.08          \\ \hline
MSFCN-2 (shared) & 88.88          & 97.08          & 93.14          & 93.58          & \textbf{86.81} & 47.47          & 75.11         & 46.78          & \textbf{88.22} & 0.27           & \textbf{32.12} & 2.27          & 95.26          \\ \hline
RFCN-2 (shared)  & 88.16          & 96.85          & 91.07          & \textbf{94.17} & 85.62          & 28.29          & \textbf{83.2} & \textbf{47.28} & 87.6           & \textbf{19.12} & 16.89          & \textbf{3.01} & 93.97          \\ \hline
MSFCN-2              & \textbf{90.01} & \textbf{97.34} & \textbf{95.97} & 93.14          & 86.76          & 73.52          & 73.63         & 35.02          & 87.86          & 3.62           & 27.57          & 1.11          & \textbf{95.35} \\ \hline
RFCN-2               & 89.48          & 97.15          & 94.01          & 93.76          & 85.88          & \textbf{76.26} & 70.35         & 39.86          & 87.5           & 8.16           & 28.05          & 1.28          & 94.67          \\ \hline
\end{tabular}
\end{table*}

\noindent \textbf{Multi-stream recurrent architecture:}
A recurrent fully convolutional network (RFCN) is designed to incorporate a recurrent network into a convolutional encoder-decoder architecture. It is illustrated in Fig \ref{fig:d}. We use the generic recurrent unit LSTM which can specialize to simpler RNNs and GRUs. LSTM operates over the encoder of previous N frames and produces a filtered encoder of the same dimension which is then fed to the decoder. \\

\noindent \textbf{Weight sharing across encoders:}
The generic form of multi-stream architectures have different weights for the different encoders. In Fig \ref{fig:video-extn} (e), the three encoders can be different and they have to be recomputed each frame. Thus the computational complexity of the encoder increases by a factor of three. However, if the weights are shared between the encoders, there is no need of recomputing it each frame. One encoder feature extraction per frame suffices and the fused encoder is computed by combination of previously computed encoders. This weight sharing approach drastically brings down the complexity with negligible additional computation relative to the single stream encoder. We demonstrate experimentally that the weight shared encoder can still provide a significant improvement in accuracy.

\section{Experiments} \label{experiments}
In this section, we explain the experimental setting including the datasets used, training algorithm details, etc and discuss the results. 

\subsection{Experimental Setup}

In most datasets, the frames in a video sequence are sparsely sampled temporally to have better diversity of objects. Thus consecutive video frames are not provided for training our multi-stream algorithm. Synthetic datasets have no cost for annotation and ground truth annotation is available for all consecutive frames. Hence we made use of the synthetic autonomous driving dataset SYNTHIA \cite{Ros_2016_CVPR} for our experiments. We also made use of DAVIS2017 \cite{Pont-Tuset_arXiv_2017} and SegTrack V2 \cite{li2013video} which provides consecutive frames, they are not automotive datasets but realistic. 

We implemented the different proposed multi-stream architectures using Keras \cite{chollet2015keras}. We used ADAM optimizer as it provided faster convergence. The maximum order (number of consecutive frames) used in the training is three (MSFCN-3) because of limitation of memory needed for training. Categorical cross-entropy is used as loss function for the optimizer. Maximum number of training epochs is set to 30 and early stopping  with a patience of 10 epochs monitoring the gains is added. Mean class IoU and per-class IoU were used as accuracy metrics. All input images were resized to 224x384 because of memory requirements needed for multiple streams. 

\subsection{Experimental Results and Discussion}

We performed four sets of experiments summarized in four tables. Qualitative results are provided in Figure 4 for KITTI, Figure 5 for DAVIS and Figure 6 for SYNTHIA. We also provide a video sequence demonstrating qualitative results for larger set of frames. 

\textbf{Table 1:} Firstly, we wanted to evaluate different orders on multi-stream and understand the impact. We also wanted to understand the impact on high speed and medium speed scenarios. SYNTHIA dataset was used for this experiment as it had separation of various speed sequences and it was also a relatively larger dataset. Two-stream networks provided a considerable increase in accuracy compared to the baseline. MSFCN-2 provided an accuracy improvement of 8\% for Highway and 14\% for City sequence. RFCN-2 provided a slightly better accuracy relative to MSFCN-2. MSFCN-3 provided marginal improvement over MSFCN-2 and thus we did not explore higher orders.

\textbf{Table 2:} KITTI is a popular automotive dataset and thus we used it to perform experiments on this real life automated driving dataset. We reduced our experiments to MSFCN-2 and RFCN-2 but we added shared weight versions of the same. MSFCN-2 provided an accuracy improvement of 11\% and the shared weight version only lagged behind slightly.

\textbf{Table 3:} We repeated the experiments of the same networks used in Table \ref{table:Kitti} on a larger SYNTHIA sequence. MSFCN-2 provided an accuracy improvement of 6\% in Mean IoU. MSFCN-2 with shared weights lagged by 1\%. RFCN-2 versions had slightly lesser accuracy compared to its MSFCN-2 counterparts with and without weight sharing. 

\textbf{Table 4:} As most automotive semantic segmentation datasets do not provide consecutive frames for temporal models, we tested in real non-auomotive datasets namely SegTrack and DAVIS. MSFCN-3 provided an accuracy improvement of 11\% in SegTrack and 6\% in DAVIS. This demonstrates that the constructed networks provide consistent improvements in various datasets. 


\begin{figure}[ht!]
    \centering
    \includegraphics[scale=0.75]{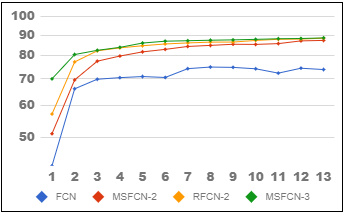}
    \caption{Accuracy over epochs for SYNTHIA dataset}
    \label{fig:epoch}
\end{figure}

\begin{table}[!htbp]
\tiny
\centering
\caption{Comparison of Multi-stream network with its baseline counterpart on DAVIS and SegTrack}
\label{table:binary}
\begin{tabular}{|l|c|c|c|c|c|}
\hline
Dataset & Architecture & Mean IoU \\ \hline
\multirow{3}{*}{ SegTrack V2} &  FCN & 83.82 \\ \cline{2-3}
                  & MSFCN-3 & \textbf{94.61} \\ \hline
\multirow{2}{*}{ DAVIS} & FCN & 77.64 \\  \cline{2-3} 
                  & MSFCN-3 & \textbf{83.42} \\  \cline{2-3}  
& BVS\cite{marki2016bilateral} & 66.52 \\ \cline{2-3} 
& FCP\cite{perazzi2015fully}  & 63.14 \\ \hline
\end{tabular}
\end{table}

\begin{figure*}[ht!]
\centering
\includegraphics[width=0.85\textwidth,height=8.5cm]{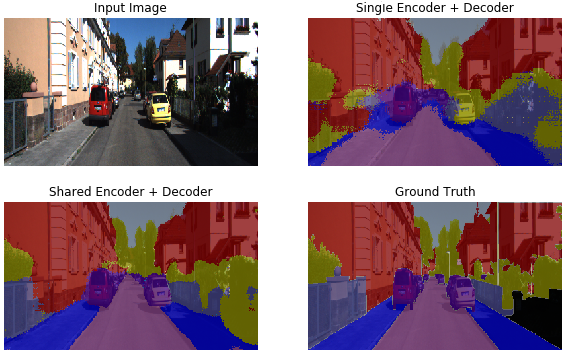}
\caption{Results on KITTI dataset}
	\label{fig:results-bin}
\end{figure*}

\begin{figure*}[ht!]
\centering
\includegraphics[width=0.85\textwidth,height=8.5cm]{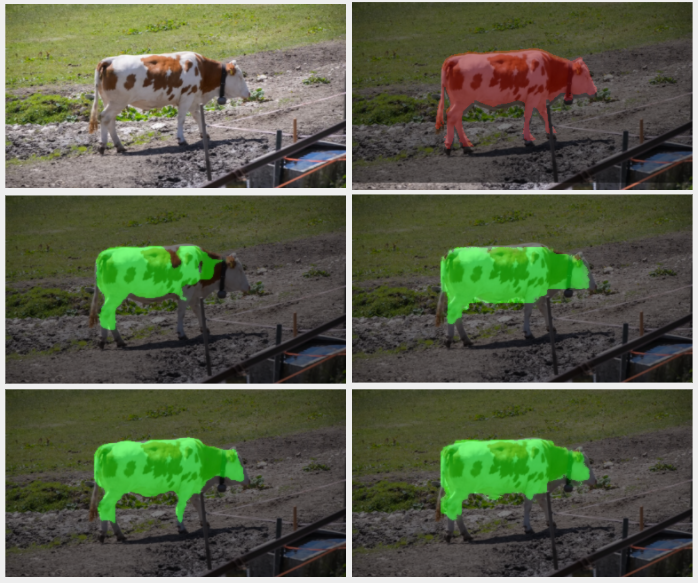}
\caption{Results over DAVIS dataset. Left to right:
RGB image, Ground Truth, Single encoder (FCN), Two stream encoder (MSFCN-2), Two stream encoder + LSTM (RFCN-2), Three stream encoder (MSFCN-3).}
	\label{fig:results-binD}
\end{figure*}

We have chosen a moderately sized based encoder namely ResNet50 and we will be experimenting with various sizes like ResNet10, ResNet101, etc for future work. In general, multi-stream provides a significant improvement in accuracy for this moderately sized encoder. The improvements might be larger for smaller networks which are less accurate. With shared weights encoder, increase in computational complexity is minimal. However, it increases memory usage and memory bandwidth quite significantly due to maintenance of additional encoder feature maps. It also increases the latency of output by 33 ms for a 30 fps video sequence. From visual inspection, the improvements are seen mainly in refining the boundaries and detecting smaller regions. It is likely due to temporal aggregation of feature maps for each pixel from past frames.

\textbf{MSFCN vs FCN:}
The single frame based FCN suffers to segment weaker classes like poles and objects at further distances. Table \ref{table:syncity} shows IoU metrics for weaker classes like Pole, Fence and Sidewalk have significantly improved in case of multi stream networks compared to single stream FCN. Fig 4 visually demonstrates that the temporal encoder modules help in preserving the small structures and boundaries in segmentation.

\textbf{MSFCN-2 vs MSFCN-3:}
The increase in the temporal information has clearly increased the performance of the semantic segmentation. But this brings an extra latency for real time applications. 

\textbf{MSFCN-2 vs RFCN:}
For a multi stream network the recurrent encoder feature fusion has shown quite a decent improvement compared to feature concatenation technique. It is also observed that the recurrent networks helped in preserving the boundaries of the weaker classes like poles and lane markings. However, RFCN demands more parameters and takes longer training time for convergence as shown in Fig \ref{fig:epoch}.

\textbf{Weight Sharing:}
In most of the experiments, MSFCN-2 with shared weights provided good improvement over the baseline and its performance deficit relative to the generic MSFCN-2 is usually small around 1\%. However, shared weights version provide a drastic improvement in computational complexity as shown by the number of parameters in Table \ref{table:Kitti}. Shared weights MSFCN-2 has a negligible increase in number of parameters and computational complexity as well whereas the generic MSFCN-2 has double the number of parameters. Thus it is important to make use of weight sharing.

\section{Conclusions} \label{conclusion}
In this paper, we designed and evaluated two video semantic segmentation architectures namely Recurrent FCN (RFCN) and Multi-Stream FCN (MSFCN) networks to exploit temporal information. We implemented three architectures namely RFCN-2, MSFCN-2 and MSFCN-3 using ResNet50 as base encoder and evaluated on SYNTHIA sequences. We obtain promising improvements of 9\% and 15\% for Highway and New York-like city scenarios over the baseline network. We also tested MSFCN-3 on real datasets like SegTrack V2 and DAVIS datasets where 11\% and 6\% accuracy improvement was achieved, respectively. We also explored weight sharing among encoders for better efficiency and produced an improvement of 11\% and 5\% for KITTI and SYNTHIA using MSFCN-2 with roughly the same complexity as the baseline encoder. In future work, we plan to explore more sophisticated encoder fusion techniques. 

\begin{figure*}[!p]
     \centering
    \includegraphics[width=\textwidth,height=17cm]{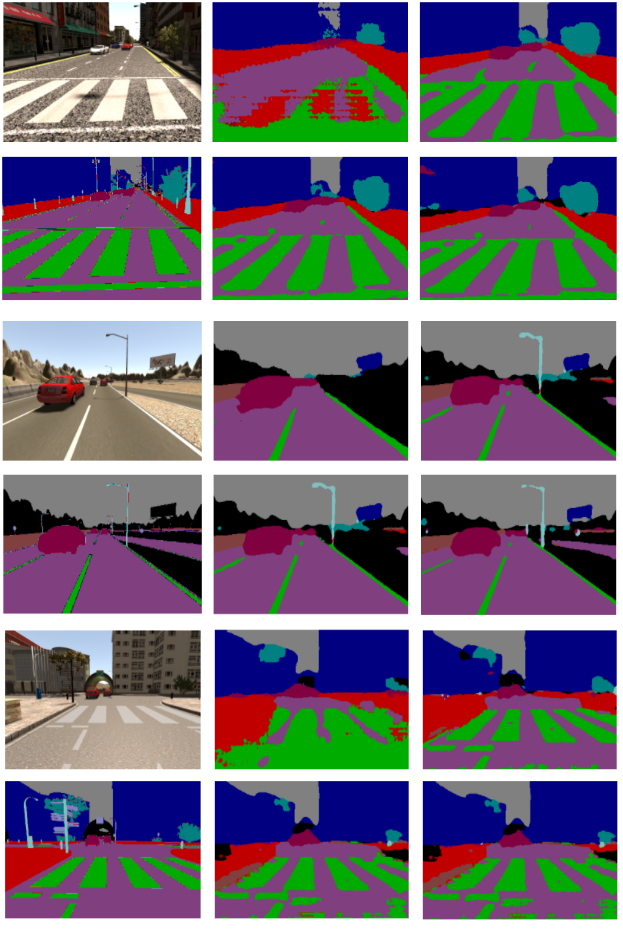}
    \caption{Qualitative results of experiments with SYNTHIA dataset. Left to right:
RGB image, Single encoder (FCN), Two stream encoder (MSFCN-2), Ground Truth, Two stream encoder + LSTM (RFCN-2) and Three stream encoder (MSFCN-3).}
    \label{fig:results}
\end{figure*}

\bibliographystyle{apalike}
{\small
\bibliography{egbib}}



\end{document}